\documentclass{article}

\PassOptionsToPackage{numbers}{natbib}
\usepackage[final]{neurips_2023}

\usepackage[utf8]{inputenc} 
\usepackage[T1]{fontenc}    
\usepackage{hyperref}       
\usepackage{url}            
\usepackage{booktabs}       
\usepackage{amsfonts}       
\usepackage{nicefrac}       
\usepackage{microtype}      
\usepackage{xcolor}         
\usepackage{soul}
\usepackage{bm}
\usepackage{graphicx} 
\usepackage{subcaption}
\usepackage[export]{adjustbox}
\usepackage{amsmath}

\title{Incorporating LLM Priors into Tabular Learners}

\author{%
 Max Zhu \\
University of Cambridge\\
mz406@cam.ac.uk\\
\And Siniša Stanivuk\\
Intellya\\
sinisa.stanivuk@intellya.ai\\
\And Andrija Petrovic\\
University of Belgrade\\
andrija.petrovic@fon.bg.ac.rs\\
\And Mladen Nikolic\\
University of Belgrade\\
mladen.nikolic@matf.bg.ac.rs\\
\And Pietro Lio\\
University of Cambridge\\
pl219@cam.ac.uk\\
}

\begin{document}

\maketitle

\begin{abstract}
We present a method to integrate Large Language Models (LLMs) and traditional tabular data classification techniques, addressing LLMs’ challenges like data serialization sensitivity and biases. We introduce two strategies utilizing LLMs for ranking categorical variables and generating priors on correlations between continuous variables and targets, enhancing performance in few-shot scenarios. We focus on Logistic Regression, introducing MonotonicLR that employs a non-linear monotonic function for mapping ordinals to cardinals while preserving LLM-determined orders. Validation against baseline models reveals the superior performance of our approach, especially in low-data scenarios, while remaining interpretable. 
\end{abstract}

\section{Introduction and related work}
Large Language Models (LLMs) have recently become the focus of research. They have been shown to contain a huge amount of world knowledge, but suffer from "hallucinating" false facts \cite{hallucination} and are sensitive to prompts \cite{step_by_step}. Concurrently, deep learning methods have struggled with tabular data \cite{tabular_data}. TabLLM \cite{tabllm} introduced large language models (LLM) and demonstrated that their world knowledge can lead to strong few-shot tabular classification performance. We propose several new methods to integrate information in LLMs in a more structured and interpretable way. Our methods are shown to improve few-shot performance against existing models as well as remaining easily interpretable.


\textbf{Pre-trained tabular learners} are models that are trained before being applied to the dataset at hand. TabPFN \cite{TabPFN} pretrains a Transformer to perform Bayesian inference on synthetic datasets, which uses in-context learning to make predictions given a small dataset. The model does not require training or parameter tuning at inference time and outperforms a variety of baselines with quick inference times. 

\textbf{LLMs as tabular models}. Tabular data is first serialized and given to a LLM as a natural language prompt and the LLM returns the classified data. This is very effective for few-shot classification on tabular datasets with semantically meaningful labels, where the LLM uses its world knowledge to predict relationships between columns and labels. In the few-shot setting, TabLLM outperforms traditional statistical and machine learning tabular classification techniques which exclusively rely on correlations within the dataset. However, TabLLM has several limitations. Firstly, fine-tuning an LLM is slow and resource expensive. Secondly, the LLMs are very sensitive to the method used for serializing tabular data into text prompts. Thirdly, the dataset's columns and labels must be understandable by the LLMs. Finally, LLMs may exhibit undesirable biases, e.g. race, sex, religion \cite{bias}. This may negatively affect predictions and, given LLMs' black-box nature, be hard to detect which further limits TabLLMs applicability in sensitive fields such as medicine or finance. Our methods solve many of these limitations by integrating LLM knowledge into existing models.

\section{Integrating LLM priors}
We propose three methods to improve the performance of existing tabular classification methods:

\textbf{Ordering categorical variables}: Categorical variables from a column are given to a LLM and the LLM sorts the categories based on how they correlate with the target attribute. For example, if a user wishes to determine a people's income, the LLM can rank the people's job descriptions. Categorical columns are augmented by using a LLM to rank categories, which are then mapped to integers and standardized (Figure \ref{fig:mono_analysis}). This is an alternative to one-hot encoding, which may give a very large input dimension if there are many categories and lead to overfitting, and mapping to ordinals based on an arbitrary order, which implies that the model must separate the categories from each other to gain useful information and may be impossible for Logistic Regression. 
Since categories are ordered in a meaningful way, a classifier can also use this ordering to extrapolate meanings between categories. 

\textbf{Priors on correlation}: The LLM uses column headers of continuous variables to predict if the column is positively or negatively correlated with the target attribute and a soft prior is applied to the classifier model. For example, the LLM can indicate that age is positively correlated with income. This is useful in the noisy or low-data regime where learning correlations can be difficult. We demonstrate this for logistic/linear regression, where we can easily apply priors. In logistic regression (LR), $\hat{y} = sigmoid(\bm{\beta} \cdot \textbf{x} + \alpha)$, a prior can be applied on $\bm{\beta}$ by minimising the training loss $\mathcal{L}$
\begin{equation}
    \mathcal{L} = BCE(y, \hat{y}) + \lambda |\bm{\beta} - \bm{\beta}_p|^2
\end{equation}
where $\hat{y}$ is model predictions, $\textbf{x}$ and $y$ are inputs and true classes, $\alpha, \bm{\beta}$, are model parameters, $\lambda$ is regularisation strength, $\bm{\beta}_p$ is the prior, and $BCE$ is binary cross-entropy loss. 
$\bm\beta$ is a vector of covariates for each column. Since input columns are standardized, entries of $\bm\beta$ will have an order of magnitude 1, so we set $\bm{\beta}_p = [-1, 0, 1]$ for the negative, absence of, or positive correlation. 

\textbf{Logistic Regression}:
In this paper, our primary focus is enhancing Logistic Regression (LR). LR is easier to interpret and apply priors to compared to neural networks and ensemble tree-based classifiers (e.g. XGBoost). The LLM can order categorical labels but does not specify their magnitudes, and LR, based on linear relations, cannot determine label magnitudes either. To remedy this, we propose MonotonicLR. It applies a learned nonlinear monotonic function to alter the weight of each category while preserving the LLM-determined order. To achieve this, a separate Unconstrained Monotonic Neural Network (UMNN) \cite{UMNN} is applied to each input column of logistic regression. For each column $i$ of the dataset, its categories $c$, $x_i^c \in \mathbb{Z} $, are mapped to a value $z_i^c \in \mathbb{R}$: 

\begin{equation} 
    \label{Eq:monotonic}
    z^c_i(x_i^c) = \int_0^{x^c_i} f(a) da, \quad \hat y = sigmoid(\bm{\beta} \cdot \textbf{z}(\textbf{x}) + \alpha)
\end{equation}

where $f$ is a neural network constrained to positive outputs with a positive activation function (Softplus \cite{Softplus} in our case). Monotonicity is guaranteed by forcing the derivative of $z$ to be positive. The sign of $\beta$ allows for increasing/decreasing functions. $z$ is integrated using the Neural ODE \cite{NODE} framework to allow parameters of $f$ to be updated with backpropagation.

Priors can be applied to MonotonicLR by regularizing $\beta_{\mathit{eff}}^i$, the average effective gradient over all categories $c$ (of the overall model) for each column $i$:

\begin{equation}
    \beta^i_{\mathit{eff}} = \sum_c \frac{\beta_i \cdot z_i^c(x^c_i)}{ x^c_i}, \quad \mathcal{L} = BCE(y, \hat{y}) + \lambda |\bm{\beta}_{\mathit{eff}} - \bm{\beta}_p| ^2
\end{equation}

In practice, we make two adjustments. Firstly, we also apply an UMNN to continuous variables, since they might need rescaling too. Secondly, we allow $f(a)$ to be slightly negative by subtracting a small bias, $f(a) = softplus(MLP(a)) - ln(2)$. This seems to help training by setting $f(0) = 0$ and allows the classifier to adapt if the LLM makes a mistake, which is easy to find since the learned mapping has a minima/maxima at the incorrect labels instead of being monotonic. See Figure \ref{fig:mono_analysis}

\section{Generating LLM priors}
LLM priors were generated using ChatGPT \cite{GPT4} during the third week of September 2023. Dataset descriptions were taken from either Kaggle or OpenML and manually serialized into either a format for column correlation or ordering categorical labels (Appendix \ref{appendix:LLM}). Despite the inability to automate the serialization due to inconsistent attribute descriptions, we followed an objective and unbiased manual approach. No prompt engineering was performed and we leave it for further exploration. 

When ordering categorical variables, ChatGPT always gave a direct ordering of categories. However, when generating priors, unless the answer gave explicit direction of correlation (e.g. "\textit{X positively correlates with Y}"), columns were marked as having no correlation with the target variable. The first response was used for all our experiments and a new chat was created for each dataset. A generic prompt and response pairs are given in Appendix \ref{appendix:LLM}. Even though we share the concern with \citet{tabllm} that ChatGPT has likely encountered these datasets during training, we believe that this method will be applicable to new unseen datasets since many attributes contain generic real-world information and are not dataset-specific. We encourage further research in this direction with new datasets. Finally, we note that all of the generated priors are easily human-interpretable. Incorrect responses/biases can easily be found and corrected, in contrast to TabLLM.

\section{Methodology and Results}
\begin{figure}
    \centering
    \includegraphics[width=0.40\linewidth]{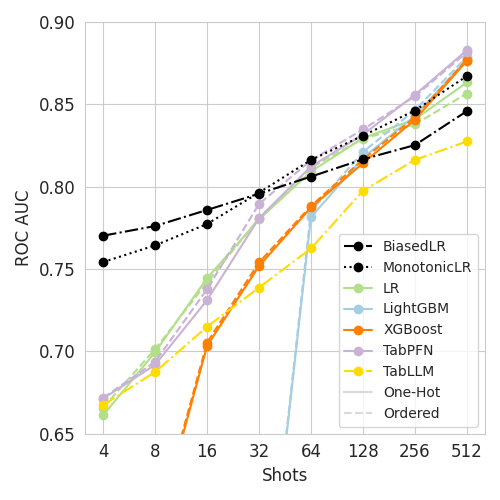}
    \includegraphics[width=0.40\linewidth]{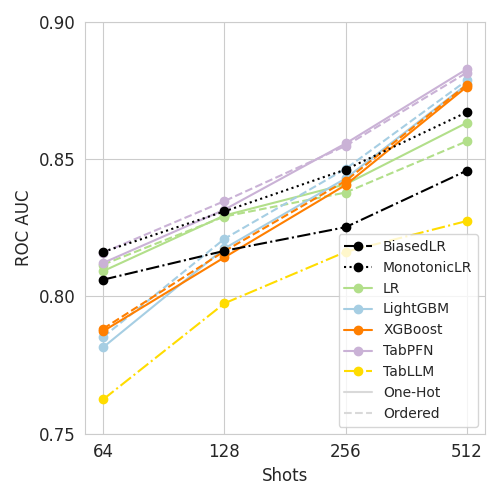}

    \caption{Area under ROC curve averaged over all datasets. Right plot is a zoomed version of the left. Shots is number of labeled rows models are trained on. }
    \label{fig:auc-results}
\end{figure}

We evaluate our method using the same few-shot setup as TabLLM \cite{tabllm}, to make our results directly comparable. We compare against baselines \textbf{TabLLM} and \textbf{TabPFN} \cite{TabPFN} as well as \textbf{LightGBM} \cite{LightGBM}, \textbf{XGBoost} \cite{XGBoost}, and Logistic Regression (\textbf{LR}). The binary classification datasets \textbf{Bank}, \textbf{California}, \textbf{CreditG}, \textbf{Diabetes}, \textbf{Income}, \textbf{Heart} and \textbf{Jungle}. Models are evaluated after fitting on subsets of the dataset with different numbers of labeled rows/shots ($n$). Baseline hyperparameters were tuned using grid search on validation tasks with the same setup as the test task, with the exception of TabLLM and TabPFN, which needs no tuning. We validated our results are within error of TabLLM's reported results so our setup is directly comparable. We take TabLLM's results directly from their paper since it requires significant computing to fine-tune and is more sensitive to the experimental setup, avoiding a chance of poor evaluation. Every baseline is evaluated on datasets with 1) \textbf{raw} data, 2) \textbf{ordered} categorical columns 3) \textbf{one-hot} encoded categorical columns (full results in Appendix \ref{FullResults}). Results are averaged over at least 20 random seeds to give around 1\% uncertainty. 

MonotonicLR and BiasedLR are fitted with the same procedure as baselines on the ordered datasets, except hyperparameters are not tuned. Tuning $\lambda$ on a validation set would leak information, $\lambda$ would depend on the quality of the LLM prior. Instead, we always scale lambda as $\lambda = 0.5 /\sqrt{n}$ for BiasedLR and $\lambda = 0.1 /\sqrt{n}$ for MonotonicLR, where $n$ is the number of shots. Priors are more strongly applied as the $n$ decreases. 

Figure \ref{fig:auc-results} shows the test area under the receiver operating characteristic curve (AUC) averaged over all datasets. Ordering the labels generally improved performance for all models versus one-hot encoding, especially TabPFN. Secondly, for a low $n$, MonotonicLR and BiasedLR strongly outperformed all baselines, demonstrating the strong impact of the LLM priors. Furthermore, both models consistently outperformed TabLLM. MonotonicLR underperforms BiasedLR at low $n$ but outperforms when more data is available. This is likely because the extra degrees of freedom leads to overfitting for small $n$. At higher $n$, our models perform more comparably to the other baselines, with MonotonicLR slightly ahead of LR. There are two possible reasons: there is enough data that the LLM prior is no longer relevant, or the underlying linear model of both BiasedLR and MonotonicLR is too simple.

\section{Analysis of MonotonicLR}
\begin{figure}
    \centering
    \begin{subfigure}[t]{0.03\textwidth}
        \textbf{a)}
    \end{subfigure}
    \begin{subfigure}{0.3\textwidth}
        \includegraphics[width=1\linewidth, valign=t]{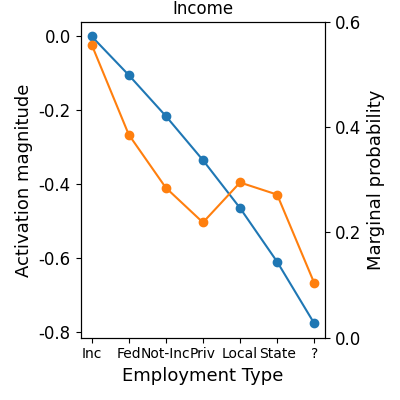}
    \end{subfigure}\hfill
    \begin{subfigure}[t]{0.03\textwidth}
        \textbf{b)}
    \end{subfigure}
    \begin{subfigure}{0.3\textwidth}
        \includegraphics[width=1\linewidth, valign=t]{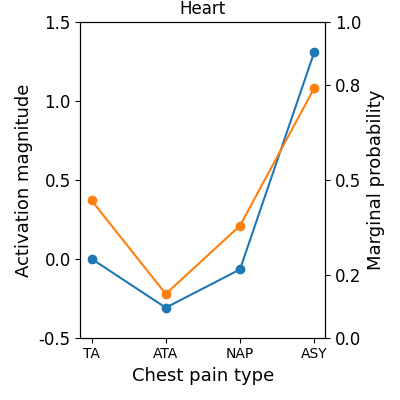}
    \end{subfigure}
    \begin{subfigure}[t]{0.015\textwidth}
        \textbf{c)}
    \end{subfigure}
    \begin{subfigure}{0.3\textwidth}
        \includegraphics[width=1\linewidth, valign=t]{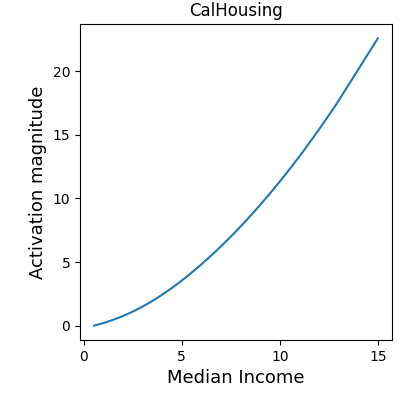}
    \end{subfigure}
        \begin{subfigure}[t]{0.015\textwidth}
        \textbf{d)}
    \end{subfigure}
    \begin{subfigure}{0.3\textwidth}
        \includegraphics[width=1\linewidth, valign=t]{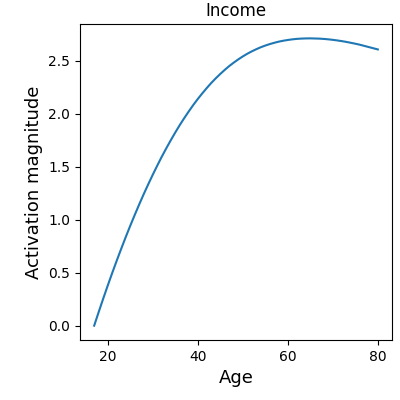}
    \end{subfigure}
    \caption{Individual variable mappings from MonotonicLR. a) and b) show model activation for categorical attributes sorted by the LLM. Blue/Left axis shows model activation, orange/right axis shows the expected outcome of entries in the category. a)/d) Employment type/age from the Income dataset. b) Chest pain type from the Heart dataset. c) Median income from the CalHousing dataset.}

    \label{fig:mono_analysis}

\end{figure}

Since MonotonicLR is based on Logistic Regression, it is easy to analyze predictions. In Equation \ref{Eq:monotonic}, we can separate out a single column $i$'s impact on the model using $\bm{\beta} \bullet \textbf{z}(\textbf{x}) = \sum{\beta_i \bullet z_i(x_i)}$. Figure \ref{fig:mono_analysis} plots out the activation magnitude $a_i = \beta_i \bullet z_i(x_i)$ for 4 different scenarios, categorical variables with correct (a) and incorrect (b) mappings generated by the LLM and continuous features that are monotonic (c) and non-monotonic (d). Higher $a_i$ means the model associates the value with the positive label. For the categorical labels, we also show the expected outcome of entries with the given category, marginalizing over all other attributes. This can be viewed as the "true" correlation and is generated with the entire dataset, while MonotonicLR is trained on a $n=512$. 

In Figure (a) the LLM orders effect of employment types on income. The ordering is correct (except for the location of "Priv" which should be lower) so the learned mapping from MonotonicLR follows a simple negative correlation. Figure (b) shows the LLM made a mistake in mapping the order of chest pain type. MonotonicLR does not give a monotonic mapping but instead has a minimum which is made possible by subtracting a small bias to the UMNN. This clearly shows the LLM has made a mistake which the MonotonicLR attempts to mitigate. 

Figure (c) shows the activation for the column Median Income in the CalHousing dataset. Our model learns a monotonic, non-linear relation between median income and house value. Figure (d) shows a U-shaped mapping for age on the Income dataset, income increases with age until around 60 when it starts decreasing. In this case, the LLM accurately predicted the "U" shaped relation (Appendix \ref{appendix:LLM}), yet we only allow for positive, negative, or no correlation as LLM priors. Integrating more complex LLM priors into models would yield better predictions, but we leave this as future work.

\section{Conclusion}
We have introduced two methods to combine LLM priors with existing tabular learning techniques, ordering categorical columns and priors on correlations, as well as MonotonicLR, an improvement over LR. In a few-shot scenario on common tabular datasets, our methods are more accurate than existing tabular classifiers. Furthermore, the LLM priors are easily interpretable and controllable. 

\section{Acknowledgements}
We would like to acknowledge GSK for supporting this work. 

\newpage
\bibliographystyle{unsrtnat}
\bibliography{references}

\begin{thebibliography}{12}
\providecommand{\natexlab}[1]{#1}
\providecommand{\url}[1]{\texttt{#1}}
\expandafter\ifx\csname urlstyle\endcsname\relax
  \providecommand{\doi}[1]{doi: #1}\else
  \providecommand{\doi}{doi: \begingroup \urlstyle{rm}\Url}\fi

\bibitem[Zhang et~al.(2023)Zhang, Li, Cui, Cai, Liu, Fu, Huang, Zhao, Zhang, Chen, Wang, Luu, Bi, Shi, and Shi]{hallucination}
Yue Zhang, Yafu Li, Leyang Cui, Deng Cai, Lemao Liu, Tingchen Fu, Xinting Huang, Enbo Zhao, Yu~Zhang, Yulong Chen, Longyue Wang, Anh~Tuan Luu, Wei Bi, Freda Shi, and Shuming Shi.
\newblock Siren's song in the ai ocean: A survey on hallucination in large language models.
\newblock 2023.

\bibitem[Kojima et~al.(2022)Kojima, Gu, Reid, Matsuo, and Iwasawa]{step_by_step}
Takeshi Kojima, Shixiang~Shane Gu, Machel Reid, Yutaka Matsuo, and Yusuke Iwasawa.
\newblock Large language models are zero-shot reasoners.
\newblock In \emph{Advances in Neural Information Processing Systems}, 2022.

\bibitem[Shwartz-Ziv and Armon(2021)]{tabular_data}
Ravid Shwartz-Ziv and Amitai Armon.
\newblock Tabular data: Deep learning is not all you need.
\newblock In \emph{8th ICML Workshop on Automated Machine Learning (AutoML)}, 2021.

\bibitem[Hegselmann et~al.(2023)Hegselmann, Buendia, Lang, Agrawal, Jiang, and Sontag]{tabllm}
Stefan Hegselmann, Alejandro Buendia, Hunter Lang, Monica Agrawal, Xiaoyi Jiang, and David Sontag.
\newblock Tabllm: Few-shot classification of tabular data with large language models.
\newblock In \emph{International Conference on Artificial Intelligence and Statistics}. PMLR, 2023.

\bibitem[Hollmann et~al.(2023)Hollmann, M{\"u}ller, Eggensperger, and Hutter]{TabPFN}
Noah Hollmann, Samuel M{\"u}ller, Katharina Eggensperger, and Frank Hutter.
\newblock Tab{PFN}: A transformer that solves small tabular classification problems in a second.
\newblock In \emph{The Eleventh International Conference on Learning Representations}, 2023.

\bibitem[Gallegos et~al.(2023)Gallegos, Rossi, Barrow, Tanjim, Kim, Dernoncourt, Yu, Zhang, and Ahmed]{bias}
Isabel~O. Gallegos, Ryan~A. Rossi, Joe Barrow, Md~Mehrab Tanjim, Sungchul Kim, Franck Dernoncourt, Tong Yu, Ruiyi Zhang, and Nesreen~K. Ahmed.
\newblock Bias and fairness in large language models: A survey.
\newblock 2023.

\bibitem[Wehenkel and Louppe(2019)]{UMNN}
Antoine Wehenkel and Gilles Louppe.
\newblock Unconstrained monotonic neural networks.
\newblock In \emph{Advances in Neural Information Processing Systems}, volume~32, 2019.

\bibitem[Glorot et~al.(2011)Glorot, Bordes, and Bengio]{Softplus}
Xavier Glorot, Antoine Bordes, and Yoshua Bengio.
\newblock Deep sparse rectifier neural networks.
\newblock In \emph{Proceedings of the Fourteenth International Conference on Artificial Intelligence and Statistics}, Proceedings of Machine Learning Research. PMLR, 2011.

\bibitem[Chen et~al.(2018)Chen, Rubanova, Bettencourt, and Duvenaud]{NODE}
Ricky T.~Q. Chen, Yulia Rubanova, Jesse Bettencourt, and David~K Duvenaud.
\newblock Neural ordinary differential equations.
\newblock In \emph{Advances in Neural Information Processing Systems}, 2018.

\bibitem[OpenAI(2023)]{GPT4}
OpenAI.
\newblock Gpt-4 technical report.
\newblock 2023.

\bibitem[Ke et~al.(2017)Ke, Meng, Finley, Wang, Chen, Ma, Ye, and Liu]{LightGBM}
Guolin Ke, Qi~Meng, Thomas Finley, Taifeng Wang, Wei Chen, Weidong Ma, Qiwei Ye, and Tie-Yan Liu.
\newblock Lightgbm: A highly efficient gradient boosting decision tree.
\newblock In \emph{Advances in Neural Information Processing Systems}, 2017.

\bibitem[Chen and Guestrin(2016)]{XGBoost}
T.~Chen and C.~Guestrin.
\newblock Xgboost: A scalable tree boosting system.
\newblock In \emph{Proceedings of the 22nd ACM SIGKDD International Conference on Knowledge Discovery and Data Mining}, 2016.

\end{thebibliography}

\newpage
\section{Appendix}
\subsection{ChatGPT prompt and response} \label{appendix:LLM}
From the dataset descriptions, we manually extract the:  \{\textit{goal of dataset}\}, \{\textit{label description(s)}\} (positive class if binary, all if multi-class), one or more \{\textit{dataset domain(s)}\}, \{\textit{column description}\}, \{\textit{target column description}\}, and for categorical variables, \{\textit{List of categories}\}. 

The prompt used for priors on column coefficient: 

\begin{quote}
    I'm creating a system to \{\textit{goal of dataset}\}. There are many factors that determine if \{\textit{label description(s)}\}, but I am interested in averaging over the unknown factors. Keep your answers short. Based on your domain knowledge of \{\textit{dataset domain(s)}\}, does \{\textit{column description}\} positively or negatively correlate with the probability of \{\textit{target column description}\}?

\end{quote}

The prompt used for ordering categorical columns:
\begin{quote}
    I'm creating an system to \{\textit{goal of dataset}\}.  I need your help to determine if a car is in acceptable state or not. There are many factors that determine if \{\textit{label description(s)}\}, but I am interested in averaging over the unknown factors. Keep your answers short. Based on your domain knowledge of \{\textit{dataset domain(s)}\}, rank the following \{\textit{column description}\} by how likely they are to influence \{\textit{target column description}\}:
    
    \{\textit{List of categories}\}
    
\end{quote}

An example of the prompt given to ChatGPT when querying for thhe continuous column \textit{Glucose} of the \textbf{Diabetes} dataset:
\begin{quote}
    I'm creating an system to diagnostically predict whether or not a patient has diabetes, based on certain diagnostic measurements included in the dataset.  There are many factors that determine if the person will have diabetes, but I am interested in averaging over the unknown factors. Keep your answers short. Based on your domain knowledge of medicine, does a person's level of plasma glucose concentration 2 hours after an oral glucose tolerance test positively or negatively correlate with the probability of that person having diabetes?
\end{quote}
And the response: 
\begin{quote}
    A person's level of plasma glucose concentration 2 hours after an oral glucose tolerance test is positively correlated with the probability of having diabetes. Higher glucose levels at this point often indicate impaired glucose tolerance, a risk factor for diabetes.
\end{quote}
This response clearly indicates a positive correlation, so we take the prior given as a positive correlation, $\bm{\beta}_p = 1$.

An example of prompt given to ChatGPT when querying for the categorical column \textit{ChestPainType} of the \textbf{Heart} dataset with categories \{\textit{TA, ATA, NAP, ASY}\}:
\begin{quote}
    I'm creating an early detection system for detecting heart failure.  I need your help to determine if a patient will die of heart failure based on their behavior. There are many factors that determine if a person will experience heart failure, but I am interested in averaging over the unknown factors. Keep your answers short. Based on your domain knowledge of medicine, rank the following chest pain types by how likely they are to influence heart failure:\\
    Typical Angina, Atypical Angina, Non-Anginal Pain, Asymptomatic
\end{quote}
And the response:
\begin{quote}
    Ranked by their likelihood to influence heart failure:\\

    1. \textbf{Typical Angina}: This type of chest pain is highly likely to be associated with heart disease and can be a significant indicator of heart failure risk.\\

    2. \textbf{Atypical Angina}: While less specific than typical angina, atypical angina can still be related to heart problems and should be considered in assessing heart failure risk.\\

    3. \textbf{Non-Anginal Pain}: Non-anginal chest pain is less likely to be directly linked to heart issues, but it should not be completely disregarded as it can sometimes be associated with heart problems.\\

    4. \textbf{Asymptomatic}: Asymptomatic individuals do not exhibit chest pain or other symptoms related to heart issues, so they are less likely to influence heart failure detection solely based on behavior.\\

    It's important to note that chest pain alone may not provide a comprehensive assessment of heart failure risk. Other risk factors, clinical evaluations, and diagnostic tests should be considered in conjunction with chest pain symptoms to make an accurate assessment.
\end{quote}

This response clearly states that the ordering of categories should be, from most influential to least influential, \{\textit{TA, ATA, NAP, ASY}\}.

\subsection{Full Results} \label{FullResults}
Full results for all baselines on raw, ordered, and one-hot datasets along with BiasedLR and MonotonicLR are shown in Table \ref{tabAll1} and \ref{tab:All2}. Datasets Blood, Diabetes and Jungle have no categorical variables so only raw results are shown. TabLLM results are taken from \cite{tabllm}. We run over significantly more random seeds so have smaller confidence intervals than their results. 

\begin{table}[h!]
\centering
\caption{AUC results for Bank, Blood, CalHousing, and Credit-G datasets, averaged over 20 random seeds. Standard deviations are shown in subscripts.}
\label{tabAll1}
\scriptsize
\begin{tabular}{lcccccccc}
\toprule
\textbf{Bank} & \multicolumn{8}{c}{Number of Shots} \\ 
\hline
Method & 4 & 8 & 16 & 32 & 64 & 128 & 256 & 512 \\
\midrule
XGBoost - Raw &$0.50_{.00}$ &$0.57_{.02}$ &$0.70_{.01}$ &$0.76_{.01}$ &$0.80_{.01}$ &$0.84_{.00}$ &$0.87_{.00}$ &$0.89_{.00}$  \\
XGBoost - Ordered &$0.50_{.00}$ &$0.57_{.02}$ &$0.70_{.01}$ &$0.77_{.01}$ &$0.82_{.01}$ &$0.86_{.00}$ &$\mathbf{0.88_{.00}} $&$\mathbf{0.90_{.00}} $ \\
XGBoost - Onehot &$0.50_{.00}$ &$0.56_{.02}$ &$0.71_{.01}$ &$0.77_{.01}$ &$0.81_{.01}$ &$0.85_{.00}$ &$0.87_{.00}$ &$\mathbf{0.90_{.00}} $ \\
LightGBM - Raw &$0.50_{.00}$ &$0.50_{.00}$ &$0.50_{.00}$ &$0.50_{.00}$ &$0.78_{.01}$ &$0.84_{.00}$ &$0.87_{.00}$ &$0.89_{.00}$  \\
LightGBM - Ordered &$0.50_{.00}$ &$0.50_{.00}$ &$0.50_{.00}$ &$0.50_{.00}$ &$0.79_{.01}$ &$0.84_{.00}$ &$\mathbf{0.88_{.00}} $&$\mathbf{0.90_{.00}} $ \\
LightGBM - Onehot &$0.50_{.00}$ &$0.50_{.00}$ &$0.50_{.00}$ &$0.50_{.00}$ &$0.79_{.01}$ &$0.83_{.00}$ &$0.87_{.00}$ &$0.89_{.00}$  \\
LR - Raw &$0.64_{.02}$ &$0.68_{.02}$ &$0.73_{.01}$ &$0.77_{.01}$ &$0.81_{.01}$ &$0.84_{.00}$ &$0.86_{.00}$ &$0.87_{.00}$  \\
LR - Ordered &$0.66_{.02}$ &$0.70_{.02}$ &$0.75_{.01}$ &$0.80_{.01}$ &$0.83_{.01}$ &$0.86_{.00}$ &$0.87_{.00}$ &$0.88_{.00}$  \\
LR - Onehot &$0.67_{.02}$ &$0.72_{.02}$ &$0.77_{.01}$ &$0.80_{.01}$ &$0.83_{.01}$ &$0.86_{.00}$ &$\mathbf{0.88_{.00}} $&$0.89_{.00}$  \\
TabPFN - Raw &$0.61_{.02}$ &$0.62_{.03}$ &$0.75_{.01}$ &$0.81_{.01}$ &$0.83_{.00}$ &$0.85_{.00}$ &$0.87_{.00}$ &$0.88_{.00}$  \\
TabPFN - Ordered &$0.63_{.02}$ &$0.62_{.02}$ &$0.76_{.01}$ &$\mathbf{0.82_{.01}} $&$\mathbf{0.85_{.00}} $&$\mathbf{0.87_{.00}} $&$\mathbf{0.88_{.00}} $&$\mathbf{0.90_{.00}} $ \\
TabPFN - Onehot &$0.64_{.02}$ &$0.60_{.02}$ &$0.72_{.01}$ &$0.78_{.01}$ &$0.82_{.00}$ &$0.85_{.00}$ &$\mathbf{0.88_{.00}} $&$\mathbf{0.90_{.00}} $ \\
TabLLM &$0.59_{.10}$ &$0.64_{.05}$ &$0.65_{.05}$ &$0.64_{.06}$ &$0.69_{.03}$ &$0.82_{.05}$ &$0.87_{.01}$ &$0.88_{.01}$  \\
BiasedLR &$\mathbf{0.78_{.00}} $&$\mathbf{0.79_{.00}} $&$\mathbf{0.80_{.00}} $&$0.81_{.00}$ &$0.83_{.00}$ &$0.84_{.00}$ &$0.85_{.00}$ &$0.87_{.00}$  \\
MonotonicLR &$0.76_{.01}$ &$0.77_{.01}$ &$0.79_{.01}$ &$0.81_{.01}$ &$0.84_{.01}$ &$0.86_{.00}$ &$\mathbf{0.88_{.00}} $&$\mathbf{0.90_{.00}} $ \\

\bottomrule
\textbf{Blood} & \multicolumn{8}{c}{Number of Shots} \\ 
\hline
Method & 4 & 8 & 16 & 32 & 64 & 128 & 256 & 512 \\
\midrule
XGBoost - Raw &$0.50_{.00}$ &$0.58_{.02}$ &$0.64_{.02}$ &$0.66_{.01}$ &$0.68_{.01}$ &$0.69_{.01}$ &$0.72_{.01}$ &-  \\
XGBoost - Ordered &$0.50_{.00}$ &$0.58_{.02}$ &$0.64_{.02}$ &$0.66_{.01}$ &$0.68_{.01}$ &$0.69_{.01}$ &$0.72_{.01}$ &-  \\
XGBoost - Onehot &$0.50_{.00}$ &$0.58_{.02}$ &$0.64_{.02}$ &$0.66_{.01}$ &$0.68_{.01}$ &$0.69_{.01}$ &$0.72_{.01}$ &-  \\
LightGBM - Raw &$0.50_{.00}$ &$0.50_{.00}$ &$0.50_{.00}$ &$0.50_{.00}$ &$0.70_{.01}$ &$0.72_{.00}$ &$0.73_{.01}$ &-  \\
LightGBM - Ordered &$0.50_{.00}$ &$0.50_{.00}$ &$0.50_{.00}$ &$0.50_{.00}$ &$0.70_{.01}$ &$0.72_{.00}$ &$0.73_{.01}$ &-  \\
LightGBM - Onehot &$0.50_{.00}$ &$0.50_{.00}$ &$0.50_{.00}$ &$0.50_{.00}$ &$0.70_{.01}$ &$0.72_{.00}$ &$0.73_{.01}$ &-  \\
LR - Raw &$0.59_{.03}$ &$0.66_{.02}$ &$0.68_{.02}$ &$\mathbf{0.72_{.01}} $&$0.73_{.00}$ &$\mathbf{0.74_{.00}} $&$0.75_{.01}$ &-  \\
LR - Ordered &$0.59_{.03}$ &$0.66_{.02}$ &$0.68_{.02}$ &$\mathbf{0.72_{.01}} $&$0.73_{.00}$ &$\mathbf{0.74_{.00}} $&$0.75_{.01}$ &-  \\
LR - Onehot &$0.59_{.03}$ &$0.66_{.02}$ &$0.68_{.02}$ &$\mathbf{0.72_{.01}} $&$0.73_{.00}$ &$\mathbf{0.74_{.00}} $&$0.75_{.01}$ &-  \\
TabPFN - Raw &$0.63_{.02}$ &$0.64_{.02}$ &$0.64_{.02}$ &$\mathbf{0.72_{.01}} $&$\mathbf{0.74_{.01}} $&$\mathbf{0.74_{.00}} $&$\mathbf{0.77_{.01}} $&-  \\
TabPFN - Ordered &$0.63_{.02}$ &$0.64_{.02}$ &$0.64_{.02}$ &$\mathbf{0.72_{.01}} $&$\mathbf{0.74_{.01}} $&$\mathbf{0.74_{.00}} $&$\mathbf{0.77_{.01}} $&-  \\
TabPFN - Onehot &$0.63_{.02}$ &$0.64_{.02}$ &$0.64_{.02}$ &$\mathbf{0.72_{.01}} $&$\mathbf{0.74_{.01}} $&$\mathbf{0.74_{.00}} $&$\mathbf{0.77_{.01}} $&-  \\
TabLLM &$0.58_{.09}$ &$0.66_{.03}$ &$0.66_{.07}$ &$0.68_{.04}$ &$0.68_{.04}$ &$0.68_{.06}$ &$0.70_{.08}$ &- \\
BiasedLR &$\mathbf{0.67_{.01}} $&$\mathbf{0.68_{.01}} $&$\mathbf{0.70_{.01}} $&$0.71_{.01}$ &$0.72_{.00}$ &$0.73_{.00}$ &$0.73_{.01}$ &-  \\
MonotonicLR &$0.64_{.02}$ &$\mathbf{0.68_{.01}} $&$0.68_{.01}$ &$\mathbf{0.72_{.01}} $&$0.73_{.01}$ &$0.73_{.00}$ &$0.74_{.01}$ &-  \\

\bottomrule
\textbf{CalHousing} & \multicolumn{8}{c}{Number of Shots} \\ 
\hline
Method & 4 & 8 & 16 & 32 & 64 & 128 & 256 & 512 \\
\midrule
XGBoost - Raw &$0.50_{.00}$ &$0.57_{.02}$ &$0.72_{.02}$ &$0.79_{.01}$ &$0.84_{.00}$ &$0.87_{.00}$ &$0.90_{.00}$ &$0.92_{.00}$  \\
XGBoost - Ordered &$0.50_{.00}$ &$0.58_{.02}$ &$0.74_{.02}$ &$0.81_{.01}$ &$0.86_{.00}$ &$0.88_{.00}$ &$0.90_{.00}$ &$0.92_{.00}$  \\
XGBoost - Onehot &$0.50_{.00}$ &$0.58_{.02}$ &$0.74_{.02}$ &$0.81_{.01}$ &$0.85_{.00}$ &$0.88_{.00}$ &$0.90_{.00}$ &$0.92_{.00}$  \\
LightGBM - Raw &$0.50_{.00}$ &$0.50_{.00}$ &$0.50_{.00}$ &$0.50_{.00}$ &$0.81_{.01}$ &$0.87_{.00}$ &$0.90_{.00}$ &$0.92_{.00}$  \\
LightGBM - Ordered &$0.50_{.00}$ &$0.50_{.00}$ &$0.50_{.00}$ &$0.50_{.00}$ &$0.83_{.01}$ &$0.88_{.00}$ &$0.90_{.00}$ &$0.92_{.00}$  \\
LightGBM - Onehot &$0.50_{.00}$ &$0.50_{.00}$ &$0.50_{.00}$ &$0.50_{.00}$ &$0.83_{.01}$ &$0.88_{.00}$ &$0.90_{.00}$ &$0.92_{.00}$  \\
LR - Raw &$0.64_{.02}$ &$0.68_{.02}$ &$0.78_{.02}$ &$0.82_{.01}$ &$0.88_{.00}$ &$0.90_{.00}$ &$0.91_{.00}$ &$0.91_{.00}$  \\
LR - Ordered &$0.68_{.02}$ &$0.71_{.02}$ &$0.79_{.01}$ &$0.83_{.01}$ &$0.88_{.00}$ &$0.90_{.00}$ &$0.91_{.00}$ &$0.91_{.00}$  \\
LR - Onehot &$0.68_{.02}$ &$0.72_{.02}$ &$\mathbf{0.81_{.01}} $&$\mathbf{0.86_{.01}} $&$\mathbf{0.89_{.00}} $&$0.90_{.00}$ &$0.91_{.00}$ &$0.92_{.00}$  \\
TabPFN - Raw &$0.66_{.02}$ &$0.67_{.01}$ &$0.77_{.02}$ &$0.83_{.01}$ &$0.88_{.01}$ &$0.90_{.00}$ &$\mathbf{0.92_{.00}} $&$\mathbf{0.93_{.00}} $ \\
TabPFN - Ordered &$0.69_{.02}$ &$0.72_{.01}$ &$0.79_{.02}$ &$0.84_{.01}$ &$\mathbf{0.89_{.00}} $&$\mathbf{0.91_{.00}} $&$\mathbf{0.92_{.00}} $&$\mathbf{0.93_{.00}} $ \\
TabPFN - Onehot &$0.68_{.03}$ &$0.71_{.02}$ &$0.79_{.02}$ &$0.84_{.01}$ &$0.88_{.00}$ &$0.90_{.00}$ &$\mathbf{0.92_{.00}} $&$\mathbf{0.93_{.00}} $ \\
TabLLM &$0.63_{.05}$ &$0.60_{.07}$ &$0.70_{.08}$ &$0.77_{.08}$ &$0.77_{.04}$ &$0.81_{.02}$ &$0.83_{.01}$ &$0.86_{.02}$  \\
BiasedLR &$\mathbf{0.76_{.01}} $&$\mathbf{0.78_{.01}} $&$0.80_{.01}$ &$0.82_{.01}$ &$0.85_{.00}$ &$0.86_{.00}$ &$0.87_{.00}$ &$0.88_{.00}$  \\
MonotonicLR &$\mathbf{0.76_{.01}} $&$\mathbf{0.78_{.01}} $&$\mathbf{0.81_{.01}} $&$0.84_{.01}$ &$0.87_{.00}$ &$0.89_{.00}$ &$0.91_{.00}$ &$0.91_{.00}$  \\

\bottomrule
\textbf{Credit-G} & \multicolumn{8}{c}{Number of Shots} \\ 
\hline
Method & 4 & 8 & 16 & 32 & 64 & 128 & 256 & 512 \\
\midrule
XGBoost - Raw &$0.50_{.00}$ &$0.55_{.02}$ &$0.59_{.01}$ &$0.64_{.01}$ &$0.68_{.01}$ &$0.71_{.01}$ &$0.74_{.00}$ &$0.76_{.01}$  \\
XGBoost - Ordered &$0.50_{.00}$ &$0.55_{.02}$ &$0.60_{.01}$ &$0.65_{.01}$ &$0.68_{.01}$ &$0.72_{.01}$ &$0.74_{.00}$ &$0.77_{.01}$  \\
XGBoost - Onehot &$0.50_{.00}$ &$0.55_{.02}$ &$0.58_{.01}$ &$0.64_{.01}$ &$0.68_{.01}$ &$0.71_{.00}$ &$0.75_{.00}$ &$0.77_{.01}$  \\
LightGBM - Raw &$0.50_{.00}$ &$0.50_{.00}$ &$0.50_{.00}$ &$0.50_{.00}$ &$0.68_{.01}$ &$0.73_{.00}$ &$0.74_{.00}$ &$0.77_{.01}$  \\
LightGBM - Ordered &$0.50_{.00}$ &$0.50_{.00}$ &$0.50_{.00}$ &$0.50_{.00}$ &$0.69_{.01}$ &$0.73_{.00}$ &$0.75_{.00}$ &$0.77_{.01}$  \\
LightGBM - Onehot &$0.50_{.00}$ &$0.50_{.00}$ &$0.50_{.00}$ &$0.50_{.00}$ &$0.66_{.01}$ &$0.71_{.00}$ &$0.74_{.00}$ &$0.77_{.01}$  \\
LR - Raw &$0.56_{.01}$ &$0.58_{.01}$ &$0.60_{.01}$ &$0.63_{.01}$ &$0.66_{.01}$ &$0.70_{.00}$ &$0.72_{.00}$ &$0.74_{.01}$  \\
LR - Ordered &$0.57_{.02}$ &$0.61_{.02}$ &$0.62_{.01}$ &$0.68_{.01}$ &$0.69_{.01}$ &$\mathbf{0.74_{.00}} $&$0.75_{.00}$ &$0.76_{.01}$  \\
LR - Onehot &$0.56_{.01}$ &$0.59_{.01}$ &$0.61_{.01}$ &$0.66_{.01}$ &$0.69_{.01}$ &$\mathbf{0.74_{.00}} $&$\mathbf{0.77_{.00}} $&$\mathbf{0.79_{.01}} $ \\
TabPFN - Raw &$0.55_{.01}$ &$0.56_{.02}$ &$0.60_{.01}$ &$0.64_{.01}$ &$0.67_{.01}$ &$0.71_{.00}$ &$0.73_{.00}$ &$0.75_{.01}$  \\
TabPFN - Ordered &$0.60_{.01}$ &$0.57_{.02}$ &$0.62_{.01}$ &$0.67_{.01}$ &$0.70_{.01}$ &$0.73_{.01}$ &$0.76_{.00}$ &$0.76_{.01}$  \\
TabPFN - Onehot &$0.59_{.01}$ &$0.58_{.02}$ &$0.63_{.01}$ &$0.68_{.01}$ &$0.71_{.01}$ &$\mathbf{0.74_{.00}} $&$\mathbf{0.77_{.00}} $&$\mathbf{0.79_{.01}} $ \\
TabLLM &$0.69_{.04}$ &$0.66_{.04}$ &$0.66_{.05}$ &$\mathbf{0.72_{.06}} $&$0.70_{.07}$ &$0.71_{.07}$ &$0.72_{.03}$ &$0.72_{.02}$  \\
BiasedLR &$\mathbf{0.71_{.00}} $&$\mathbf{0.71_{.00}} $&$\mathbf{0.71_{.00}} $&$\mathbf{0.72_{.00}} $&$\mathbf{0.72_{.00}} $&$0.73_{.00}$ &$0.74_{.00}$ &$0.75_{.01}$  \\
MonotonicLR &$0.70_{.01}$ &$0.69_{.01}$ &$0.68_{.01}$ &$0.70_{.01}$ &$0.70_{.01}$ &$0.73_{.00}$ &$0.75_{.00}$ &$0.76_{.01}$  \\

\bottomrule

\end{tabular}
\end{table}

\begin{table}[h!]
\caption{AUC results for Diabetes, Heart, Income, and Jungle datasets, averaged over 20 random seeds. Standard deviations are shown in subscripts. }
\label{tab:All2}
\scriptsize
\centering
\begin{tabular}{lcccccccc}
\toprule
\textbf{Diabetes} & \multicolumn{8}{c}{Number of Shots} \\ 
\hline
Method & 4 & 8 & 16 & 32 & 64 & 128 & 256 & 512 \\
\midrule
XGBoost - Raw &$0.50_{.00}$ &$0.60_{.02}$ &$0.71_{.01}$ &$0.74_{.01}$ &$0.77_{.01}$ &$0.79_{.00}$ &$0.81_{.00}$ &$0.83_{.01}$  \\
LightGBM - Raw &$0.50_{.00}$ &$0.50_{.00}$ &$0.50_{.00}$ &$0.50_{.00}$ &$0.78_{.00}$ &$0.80_{.00}$ &$0.83_{.00}$ &$0.83_{.01}$  \\
LR - Raw &$0.65_{.03}$ &$0.68_{.02}$ &$0.73_{.01}$ &$0.77_{.01}$ &$0.80_{.00}$ &$0.81_{.00}$ &$0.83_{.00}$ &$0.83_{.01}$  \\
TabPFN - Raw &$0.68_{.01}$ &$0.68_{.03}$ &$0.68_{.02}$ &$0.76_{.01}$ &$0.80_{.00}$ &$0.81_{.01}$ &$\mathbf{0.84_{.00}} $&$\mathbf{0.86_{.01}} $ \\
TabLLM &$0.61_{.09}$ &$0.63_{.08}$ &$0.69_{.07}$ &$0.68_{.04}$ &$0.73_{.03}$ &$0.79_{.04}$ &$0.78_{.02}$ &$0.78_{.04}$  \\
BiasedLR &$\mathbf{0.80_{.00}} $&$\mathbf{0.79_{.00}} $&$\mathbf{0.80_{.00}} $&$\mathbf{0.80_{.00}} $&$\mathbf{0.81_{.00}} $&$\mathbf{0.82_{.00}} $&$0.82_{.00}$ &$0.84_{.01}$  \\
MonotonicLR &$0.77_{.01}$ &$0.76_{.01}$ &$0.77_{.01}$ &$0.79_{.00}$ &$\mathbf{0.81_{.00}} $&$\mathbf{0.82_{.00}} $&$0.83_{.00}$ &$0.84_{.01}$  \\
\bottomrule

\textbf{Heart} & \multicolumn{8}{c}{Number of Shots} \\ 
\hline
Method & 4 & 8 & 16 & 32 & 64 & 128 & 256 & 512 \\
\midrule
XGBoost - Raw &$0.50_{.00}$ &$0.65_{.03}$ &$0.81_{.02}$ &$0.85_{.01}$ &$0.88_{.00}$ &$0.90_{.00}$ &$0.92_{.00}$ &$\mathbf{0.93_{.00}} $ \\
XGBoost - Ordered &$0.50_{.00}$ &$0.65_{.03}$ &$0.81_{.01}$ &$0.85_{.01}$ &$0.88_{.00}$ &$0.90_{.00}$ &$0.92_{.00}$ &$\mathbf{0.93_{.00}} $ \\
XGBoost - Onehot &$0.50_{.00}$ &$0.65_{.03}$ &$0.81_{.01}$ &$0.85_{.01}$ &$0.88_{.00}$ &$0.90_{.00}$ &$0.92_{.00}$ &$0.92_{.00}$  \\
LightGBM - Raw &$0.50_{.00}$ &$0.50_{.00}$ &$0.50_{.00}$ &$0.50_{.00}$ &$0.88_{.00}$ &$0.91_{.00}$ &$0.92_{.00}$ &$\mathbf{0.93_{.00}} $ \\
LightGBM - Ordered &$0.50_{.00}$ &$0.50_{.00}$ &$0.50_{.00}$ &$0.50_{.00}$ &$0.88_{.00}$ &$0.91_{.00}$ &$0.92_{.00}$ &$\mathbf{0.93_{.00}} $ \\
LightGBM - Onehot &$0.50_{.00}$ &$0.50_{.00}$ &$0.50_{.00}$ &$0.50_{.00}$ &$0.89_{.00}$ &$0.91_{.00}$ &$0.92_{.00}$ &$\mathbf{0.93_{.00}} $ \\
LR - Raw &$0.79_{.03}$ &$0.83_{.02}$ &$0.85_{.02}$ &$0.86_{.01}$ &$0.90_{.00}$ &$0.91_{.00}$ &$0.91_{.00}$ &$0.91_{.00}$  \\
LR - Ordered &$0.79_{.03}$ &$0.83_{.02}$ &$0.85_{.01}$ &$0.86_{.01}$ &$0.90_{.00}$ &$0.91_{.00}$ &$0.91_{.00}$ &$0.91_{.00}$  \\
LR - Onehot &$0.79_{.03}$ &$0.83_{.02}$ &$0.86_{.01}$ &$0.87_{.01}$ &$\mathbf{0.91_{.00}} $&$0.91_{.00}$ &$0.92_{.00}$ &$0.92_{.00}$  \\
TabPFN - Raw &$0.81_{.02}$ &$0.86_{.01}$ &$\mathbf{0.89_{.00}} $&$\mathbf{0.89_{.00}} $&$\mathbf{0.91_{.00}} $&$\mathbf{0.92_{.00}} $&$0.92_{.00}$ &$\mathbf{0.93_{.00}} $ \\
TabPFN - Ordered &$0.83_{.01}$ &$0.86_{.01}$ &$\mathbf{0.89_{.00}} $&$\mathbf{0.89_{.00}} $&$\mathbf{0.91_{.00}} $&$\mathbf{0.92_{.00}} $&$0.92_{.00}$ &$\mathbf{0.93_{.00}} $ \\
TabPFN - Onehot &$\mathbf{0.85_{.01}} $&$\mathbf{0.88_{.01}} $&$\mathbf{0.89_{.00}} $&$\mathbf{0.89_{.00}} $&$\mathbf{0.91_{.00}} $&$\mathbf{0.92_{.00}} $&$\mathbf{0.93_{.00}} $&$\mathbf{0.93_{.00}} $ \\
TabLLM &$0.76_{.14}$ &$0.83_{.05}$ &$0.87_{.04}$ &$0.87_{.06}$ &$\mathbf{0.91_{.01}} $&$0.90_{.01}$ &$0.92_{.01}$ &$0.92_{.01}$  \\

BiasedLR &$0.83_{.00}$ &$0.84_{.01}$ &$0.85_{.00}$ &$0.86_{.00}$ &$0.88_{.00}$ &$0.89_{.00}$ &$0.90_{.00}$ &$0.91_{.00}$  \\
MonotonicLR &$0.84_{.01}$ &$0.86_{.01}$ &$0.87_{.01}$ &$0.88_{.00}$ &$0.90_{.00}$ &$\mathbf{0.92_{.00}} $&$\mathbf{0.93_{.00}} $&$\mathbf{0.93_{.00}} $ \\
\bottomrule

\textbf{Income} & \multicolumn{8}{c}{Number of Shots} \\ 
\hline
Method & 4 & 8 & 16 & 32 & 64 & 128 & 256 & 512 \\
\midrule
XGBoost - Raw &$0.50_{.00}$ &$0.59_{.02}$ &$0.71_{.01}$ &$0.77_{.01}$ &$0.81_{.01}$ &$0.85_{.00}$ &$0.88_{.00}$ &$\mathbf{0.90_{.00}} $ \\
XGBoost - Ordered &$0.50_{.00}$ &$0.61_{.02}$ &$0.76_{.01}$ &$0.81_{.01}$ &$0.82_{.01}$ &$0.86_{.00}$ &$\mathbf{0.89_{.00}} $&$\mathbf{0.90_{.00}} $ \\
XGBoost - Onehot &$0.50_{.00}$ &$0.59_{.02}$ &$0.74_{.01}$ &$0.79_{.01}$ &$0.82_{.01}$ &$0.86_{.00}$ &$0.88_{.00}$ &$\mathbf{0.90_{.00}} $ \\
LightGBM - Raw &$0.50_{.00}$ &$0.50_{.00}$ &$0.50_{.00}$ &$0.50_{.00}$ &$0.80_{.01}$ &$0.84_{.00}$ &$0.87_{.00}$ &$0.89_{.00}$  \\
LightGBM - Ordered &$0.50_{.00}$ &$0.50_{.00}$ &$0.50_{.00}$ &$0.50_{.00}$ &$0.83_{.01}$ &$0.86_{.00}$ &$0.88_{.00}$ &$\mathbf{0.90_{.00}} $ \\
LightGBM - Onehot &$0.50_{.00}$ &$0.50_{.00}$ &$0.50_{.00}$ &$0.50_{.00}$ &$0.83_{.01}$ &$0.85_{.00}$ &$0.87_{.00}$ &$\mathbf{0.90_{.00}} $ \\
LR - Raw &$0.66_{.02}$ &$0.69_{.02}$ &$0.74_{.01}$ &$0.76_{.01}$ &$0.81_{.00}$ &$0.82_{.00}$ &$0.83_{.00}$ &$0.85_{.00}$  \\
LR - Ordered &$0.75_{.03}$ &$0.77_{.02}$ &$0.81_{.01}$ &$0.84_{.01}$ &$\mathbf{0.87_{.00}} $&$\mathbf{0.88_{.00}} $&$\mathbf{0.89_{.00}} $&$0.89_{.00}$  \\
LR - Onehot &$0.72_{.03}$ &$0.76_{.02}$ &$0.79_{.01}$ &$0.82_{.01}$ &$0.84_{.00}$ &$0.87_{.00}$ &$0.88_{.00}$ &$0.89_{.00}$  \\
TabPFN - Raw &$0.64_{.02}$ &$0.75_{.01}$ &$0.75_{.01}$ &$0.77_{.01}$ &$0.80_{.01}$ &$0.84_{.00}$ &$0.87_{.00}$ &$0.88_{.00}$  \\
TabPFN - Ordered &$0.69_{.02}$ &$0.81_{.01}$ &$0.80_{.01}$ &$0.85_{.00}$ &$0.84_{.01}$ &$0.87_{.00}$ &$\mathbf{0.89_{.00}} $&$\mathbf{0.90_{.00}} $ \\
TabPFN - Onehot &$0.67_{.02}$ &$0.81_{.01}$ &$0.79_{.01}$ &$0.82_{.01}$ &$0.83_{.01}$ &$0.85_{.00}$ &$0.88_{.00}$ &$0.88_{.00}$  \\
TabLLM &$0.84_{.01}$ &$0.84_{.02}$ &$0.84_{.04}$ &$0.84_{.01}$ &$0.84_{.02}$ &$0.86_{.01}$ &$0.87_{.00}$ &$0.89_{.01}$  \\

BiasedLR &$\mathbf{0.86_{.00}} $&$\mathbf{0.86_{.00}} $&$\mathbf{0.86_{.00}} $&$\mathbf{0.86_{.00}} $&$\mathbf{0.87_{.00}} $&$\mathbf{0.88_{.00}} $&$0.88_{.00}$ &$0.88_{.00}$  \\
MonotonicLR &$0.85_{.01}$ &$0.84_{.01}$ &$0.85_{.01}$ &$0.85_{.01}$ &$0.86_{.00}$ &$\mathbf{0.88_{.00}} $&$\mathbf{0.89_{.00}} $&$\mathbf{0.90_{.00}} $ \\
\bottomrule

\textbf{Jungle} & \multicolumn{8}{c}{Number of Shots} \\ 
\hline
Method & 4 & 8 & 16 & 32 & 64 & 128 & 256 & 512 \\
\midrule
XGBoost - Raw &$0.50_{.00}$ &$0.58_{.02}$ &$0.69_{.01}$ &$0.75_{.01}$ &$0.80_{.01}$ &$0.83_{.01}$ &$\mathbf{0.88_{.00}} $&$\mathbf{0.91_{.00}} $ \\
LightGBM - Raw &$0.50_{.00}$ &$0.50_{.00}$ &$0.50_{.00}$ &$0.50_{.00}$ &$0.78_{.01}$ &$\mathbf{0.84_{.00}} $&$\mathbf{0.88_{.00}} $&$0.90_{.00}$  \\
LR - Raw &$0.63_{.02}$ &$0.65_{.02}$ &$0.71_{.01}$ &$0.74_{.01}$ &$0.78_{.00}$ &$0.79_{.00}$ &$0.80_{.00}$ &$0.81_{.00}$  \\
TabPFN - Raw &$0.63_{.02}$ &$0.64_{.02}$ &$0.71_{.01}$ &$0.75_{.01}$ &$0.80_{.00}$ &$0.83_{.00}$ &$0.86_{.00}$ &$0.90_{.00}$  \\
TabLLM &$0.64_{.01}$ &$0.64_{.02}$ &$0.65_{.03}$ &$0.71_{.02}$ &$0.78_{.02}$ &$0.81_{.02}$ &$0.84_{.01}$ &$0.89_{.01}$  \\

BiasedLR &$\mathbf{0.75_{.00}} $&$\mathbf{0.75_{.00}} $&$\mathbf{0.76_{.00}} $&$0.77_{.00}$ &$0.78_{.00}$ &$0.79_{.00}$ &$0.80_{.00}$ &$0.80_{.00}$  \\
MonotonicLR &$0.72_{.01}$ &$0.73_{.01}$ &$\mathbf{0.76_{.01}} $&$\mathbf{0.78_{.01}} $&$\mathbf{0.81_{.00}} $&$0.82_{.00}$ &$0.84_{.00}$ &$0.84_{.00}$  \\
\bottomrule
\end{tabular}
\end{table}

\end{document}